# Heuristic Classification of Thoughts Prompting (HCoT): Integrating Expert System Heuristics for Structured Reasoning into Large Language Models

Lei LIN[a], Jizhao ZHU[b]*,Yong LIU[c], Donghong SUN[d], Hongbo HE[a], Yihua DU[a]

*a, Computer Network Information Center, Chinese Academy of Sciences, Beijing, China*
*b, Shenyang Aerospace University*
*c, ZGC LAB, Beijing,China*
*d, Institute for Network Sciences and Cyberspace, Tsinghua University, Beijing, China*

*Lei LIN, email: linlei@cnic.cn*
*Jizhao ZHU, email: zhujz@sau.edu.cn*
*Yong LIU, email: liuyong@zgclab.edu.cn*
*Donghong SUN, email: sundonghong@tsinghua.edu.cn*
*Hongbo HE, email: hhb@cnic.cn*
*Yihua DU, email: yhdu@cashq.ac.cn*
*This work was supported by the Strategic Priority Research Program of Chinese Academy of Sciences under Grant XDA0480301*
*Present/permanent address:Computer Network Information Center, Yard 2, Dongsheng South Road, Haidian District, Beijing, China*

**Abstract**

Large language models (LLMs) suffer from two limitations when attempting to solve complex problems. First, their reasoning processes exhibit Bayesian-like stochastic generation properties, where each token is sampled from a probability distribution conditioned on the preceding context. This can lead to decision trajectories with inherent randomness rather than deterministic planning. Second, the reasoning processes and decision-making mechanisms in LLMs are statically decoupled. This means domain knowledge is dynamically retrieved but it does not dynamically alter the underlying reasoning strategy. Given that stochastic generation does not have any mechanisms for correcting trajectories or using knowledge to guide the optimization process during sequential reasoning, this double deficiency not only means initial decisions lack strategic anchoring, but also that reasoning chains often fail to converge on the correct solutions. To solve these two problems, this paper presents a problem-solving method designed to be incorporated into an LLM's generation process as a way of guiding reasoning. The method is widely compatible with numerous LLMs and features reusable solutions. The method is essentially based on a new Heuristic-Classification-of-Thoughts prompting schema (HCoT) which tries to synergize the LLM's reasoning ability with a structured problem space. A heuristic classification model controls the LLM's reasoning process, providing reusable abstract solutions. Performing HCoT on two complex inductive reasoning problems, where the search space could not be well-defined, this solution returned better results than several current problem-solving approaches, including Tree-of-Thoughts and Chain-of-Thoughts prompting. The method was also evaluated on the well-structured reasoning task of 24 Game. Compared to the state-of-the-art approach Tree-of -Thoughts-Breadth-First-Search, HCoT was significantly more token-efficient. In terms of both accuracy and token usage, this solution lies on the Pareto frontier relative to prior methods, achieving a strong balance between performance and computational cost.

Keyword: Problem Solving Method, Heuristic Classification, Large Language Model, Thought Prompting

## 1. Introduction

Large language Models (LLMs) such as GPT (Radford, 2018), DeepSeek (DeepSeek-AI, 2024; DeepSeek-AI et al., 2024), and ERNIE bot (Huang et al., 2024), have demonstrated strong abilities with a range of problem-solving tasks. In addition to traditional NLP problems, like text classification, text generation, and information extraction, LLMs are also capable of complex problem-solving abilities such as inductive reasoning (R. Wang et al., 2024), mathematical problem-solving (Shao et al., 2024), code generation and debugging (Yeo et al., 2024), and logical inference (Toroghi et al., 2024). It is perhaps surprising that at the core of this remarkable progress lies a text-generation process driven by Bayesian inference. These processes are fundamentally grounded in an autoregressive mechanism that crafts reasoning through sequential, left-to-right token-level decisions. Derived from patterns in the training data, these processes are inherently prone to stochastic outputs. Additionally, they are highly susceptible to deviations from a solidly reasoned pathway. In humans, when some logic embedded in experience diverges from real-world problem-solving contexts, we call this tendency cognitive bias. Consequently, a critical question arises: How can we develop methodologies that systematically mitigate deviations from a reasoned pathway induced by mismatches in the distributions of the training data vs. a real-world problem domain without resorting to task-specific fine-tuning or retraining?

Fortunately, the literature on problem-solving provides some clues to answer this question (Chandrasekaran, 1990; Chang et al., 2018; Clancey, 1985; Fensel & Motta, 2001; Vitalari & Dickson, 1983). Research on "design problem solving" or "structured analytic techniques" suggests that a structured problem space could improve systematic bias while reducing random noise (Chandrasekaran, 1990; Chang et al., 2018). In fact, researchers have used numerous machine-learning-based problem-solving methods for this purpose. For example, human-in-the-loop approaches try to remove noise through human evaluation and feedback during training. Since humans are generally capable of providing much more comprehensive domain knowledge and experience, this extra step helps the model better understand the true situation behind the data. Consequently, models tend to perform better and are less affected by data bias (Wu et al., 2022). The conversational nature of LLMs makes it possible to have humans guide whether to generate or remove noise. It also allows humans to add prior domain knowledge to help offset deviations from reason.

However, designing this kind of problem-solving process requires returning to the fundamentals of problem-solving. To this end, I drew inspiration from how Newell chose to explore problem spaces and task structures way back in the 1980s (Newell, 1981). Newell characterized problem-solving as the task of searching through a structured problem space. In the same vein, Chandrasekaran defines a generic task as a problem/method/knowledge/inference package (Bylander & Chandrasekaran, 1987; Chandrasekaran, 1989). These form the underlying principles of the problem-solving method for LLMs outlined in this paper.

As Fig. 1 illustrates, most of the existing problem-solving methods sample knowledge from a trained LLM. However, the Heuristic-Classification-of-Thoughts (HCoT) method was designed to "think about" the types of fixed knowledge structures and inference strategies described in Benjamins (1998) and Bylander & Chandrasekaran (1987). Establishing a problem-solving method based on thought involves combining a problem with domain experts who define an inference strategy after acquiring knowledge. As described in Fig 1(d), the inference strategy in problem solving method contains a prior defined role (thought) that is an intermediate step toward problem-solving. The role contains all inputs, outputs, and domain knowledge required by inference. This kind of programmed inference process can replace the cognitive modeling of the LLM to both reduce randomness and standardize the inference trajectory. It can also use prior knowledge about the problem-solving method and reuse existing solutions to overcome any limitations of the LLM's reasoning created during the training process. In other words, LLMs generally search for a solution in a generated problem space, as is the case with Chain-of-Thoughts and Tree-of-Thoughts. However, with a thought-based problem-solving method, the LLM searches an established problem space formed by both the model's reasoning and prior domain knowledge. The thought-based problem-solving method presented in this paper, HCoT, is based on the most classical problem-solving method – heuristic classification (Clancey, 1985; Fensel & Motta, 2001).

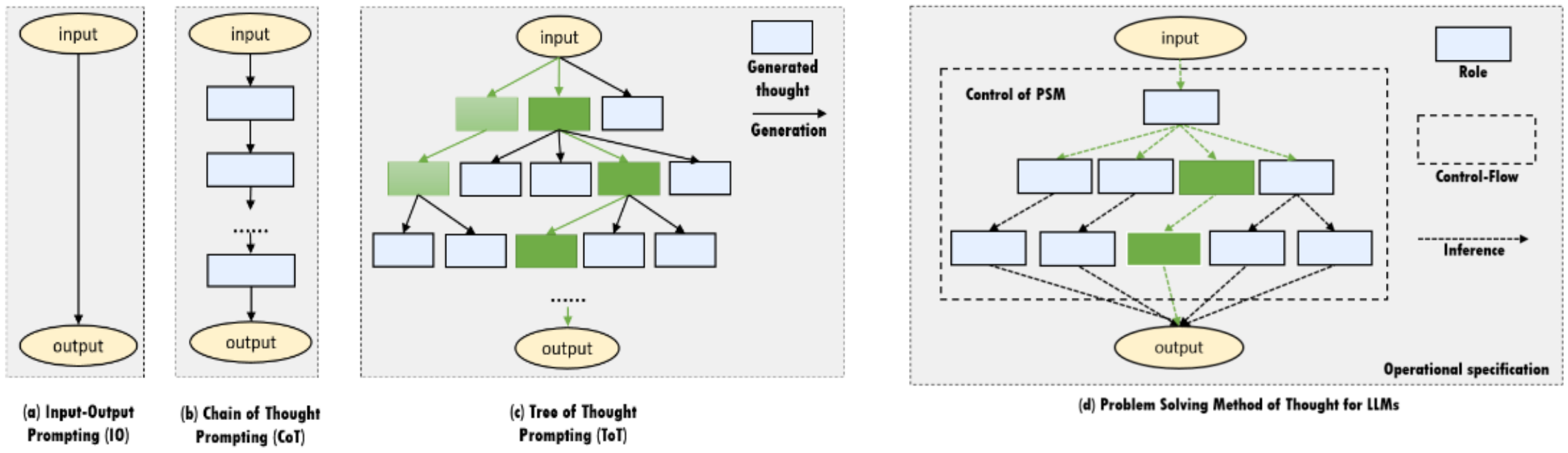

Fig. 1. Different methodologies for leveraging LLMs in problem-solving scenarios.

HCoT addresses two induction reasoning problems: the list function problem (Rule, 2020) and the 1D-ARC (R. Wang et al., 2024) problem, both of which are widely acknowledged as being complex and ill-structured problems (Reed, 2016; Simon, 1973). Further, to rigorously validate the efficacy of our approach, even with well-structured reasoning problems, comprehensive experiments were conducted using scenarios based on the 24 Game. Experiments involving two different induction tasks unequivocally demonstrate that HCoT outperforms input-output prompting, Chain-of-Thoughts prompting, and Tree-of-Thoughts prompting. Moreover, HCoT stands out as an efficient solution for the 24 Game, achieving competitive accuracy with significantly fewer generated tokens compared to methods like Tree-of-Thoughts-Breadth-First-Search (ToT-BFS) (Yao et al., 2023). As such, HCoT showcases superior reasoning efficiency. In light of these findings and the insights gained from these research endeavors, the contributions of this study are summarized as follows:

- This paper presents a problem-solving method, called HCoT, that regularizes the reasoning process of LLMs. HCoT's reasoning process is explicitly defined and formally modeled. The formalization specifies each inference step, ensuring a clear and structured logical trajectory for the model.
- HCoT is based on heuristic classification. Compared to other prompting approaches, like Tree-of-Thoughts, Chain-of-Thoughts, and Graph-of-Thoughts, HCoT not only harnesses the power of reusable reasoning patterns to boost the problem-solving capabilities of LLMs, it also pioneers a novel problem structure for solving ill-structured problems in the context of LLMs. This innovative structure acts as a beacon, illuminating the path for LLMs to tackle the more convoluted and often vague nature of ill-structured problems, opening up new avenues for more effective reasoning.
- Experiments on two inductive reasoning datasets demonstrate that HCoT not only enhances the performance of general-purpose LLMs, like DeepSeek-V3 and GPT-4o, but also yields significant improvements on models that already excel at complex problem-solving, such as DeepSeek-R1 (DeepSeek-AI, 2024) and GPT-4o. Our experiments with the 24 Game indicate that the HCoT method significantly reduces the number of generated tokens, effectively shortening the inference path while maintaining a high success rate. This highlights HCoT's ability to enhance efficiency without compromising the accuracy of the solution.

## 2. Literature Review

Over the last three years, research on problem-solving with LLMs has evolved from “enabling models to articulate their thought processes” to “empowering models to search, execute and reflect” (Wei et al., 2022)

Wei and colleagues went on to propose Chain-of-Thoughts (CoT) prompting, where LLMs imitate a reasoning path through reasoning examples provided by users. Z. Zhang et al. (2023) followed up with an automatic CoT model, which prompts LLMs to "think step by step". Consequently, the LLMs generate their own reasoning paths via a zero-shot approach. However, the selection mechanism, which is inherent to the CoT approach, is greedy and therefore represents a notable deficiency. This limitation contributes to the inherent self-owned decision bias exhibited by LLMs – a bias that stems from the latent prejudices embedded within their training data. X. Wang et al. (2023) proposed self-consistency CoT, rather than adhering solely to the greedy approach of selecting a single reasoning path. The method initially samples a broad and varied set of reasoning paths. Subsequently, it determines the most consistent answer by marginalizing answer probabilities over these sampled reasoning paths, making the reasoning outcomes more robust.

A few years later Yao et al. (2023) introduced Tree-of-Thoughts (ToT) prompting. Unlike the self-reliant CoT, which solely evaluates one final set of generated conclusions, ToT assesses and selects intermediate results at each step of the reasoning process. This equips ToT with a robust search strategy for solving well-defined problems (Russell & Norvig, 2016). For instance, ToT has achieved state-of-the-art performance on the 24 Game, demonstrating just how effectively it can optimize the reasoning pathway.

The very next year, Besta et al. (2024) proposed the Graph-of-Thoughts (GoT) method, a novel framework that enhances the prompting capabilities of LLMs by modeling the reasoning process as an arbitrary graph, rather than being constrained to tree-like structures as with the traditional paradigms like CoT or ToT. This approach lets the arbitrary thoughts generated by LLMs be merged and refined through graph-based transformations, effectively reducing dependency hierarchies in the reasoning path. GoT demonstrates significant improvements with sorting and set intersection tasks, increasing the quality of results over ToT by as much as 62% while simultaneously reducing costs by more than 31%. However, GoT was primarily designed to address well-defined problems where operations can be generated to explicitly define a problem's search space. What all these methods struggle with is solving ill-structured problems (Reed, 2016; Simon, 1973)– problems that lack a clearly constructible search space.

Solving ill-structured problems requires defining a suitable problem-solving structure (Pople, 2019). In the past, researchers have focused on heuristic approaches, some of which have seen success, particularly in medical diagnostics (Clancey, 1985; Pople, 2019). Clancey was the first to propose heuristic classification as a tool for structuring problem spaces so as to solve ill-structured problems back in 1985. This first approach was developed in the context of medical diagnosis, and then extended to personalized book recommendations, among other applications. In recent years, heuristic methods in medical diagnostics have been replaced by machine learning (Teng et al., 2022) and case-based reasoning (Mustafa et al., 2023). Unlike heuristic classification, which depends on an extensive set of general rules, case-based reasoning leverages concrete knowledge from past cases to reduce the upfront effort of acquiring knowledge. Then, by retaining new experiences incrementally, the models continually adapt and refine their solutions across a diverse set of scenarios. Notably, high-performance case-based reasoning systems can still benefit from targeted domain knowledge, especially in terms of steering which cases to retrieve and when attempting to adapt to new knowledge (Agnar & Plaza, 1994). However, although deep learning has demonstrated remarkable efficacy within the medical domain, achieving accuracy rates that are comparable to or even superior to those demonstrated by human experts has been difficult. Additionally, challenges with interpretability and explainability have hindered the broader practical implementation of deep learning in clinical practice (Teng et al., 2022).

Interestingly, the emergence of LLMs has seen the field revert back to some of the heuristic machine-learning approaches of old (Caruccio et al., 2024). In fact, one of the most important applications of an LLM in the field of medical diagnostics is to use some form of prompting to match a patient's symptoms with a potential diagnosis (Caruccio et al., 2024). This, of course, is a key step in heuristic classification (Clancey, 1985; Fensel & Motta, 2001). However, one must ask: When classifying a problem solution heuristically, can LLMs perform the whole problem-solving process? This paper presents a heuristic classification inference schema to guide LLMs in solving heuristic classification problems. Through this approach, the LLM will be able to structure the problem using prior domain knowledge on the one hand and a fixed reasoning path on the other.

## 3. Methodology

This section begins by formalizing the existing methodologies that leverage LLMs for problem-solving, with a particular focus on delineating their structural components and operational workflows. This formalization is then used to inform a structured comparison of the theoretical strengths and weaknesses of the different "of-Thought" paradigms for problem-solving. The section concludes by outlining the Heuristic-Classification-of-Thoughts (HCoT) framework as a concrete instantiation of the "of-Thought" approach. Its structure and operational mechanisms are described in great detail so as to enable rigorous theoretical analysis.

### *3.1. Theoretical Foundations*

$p_\theta$ denotes a pre-trained language model with a set of parameters $\theta$. The lowercase letters $x, y, z, \cdots$ denote a language sequence, i.e. $x = (x^1, \cdots x^n)$ where each $x^i$ is a token, $p_\theta(x) = \prod_{i=1}^{n} p_\theta(x^i | x^1, x^2, \cdots x^{i-1})$ and $x^1, x^2, \cdots x^{i-1}$ present sequence of generation.

Input-output prompting, such as zero-shot prompting and few-shot prompting (Sahoo et al., 2024), extracts a sample $y$ from the distribution $y \sim p_\theta(y|x)$ given the input prompt $x$, where $x$ is a task instruction or a few-shot example.

CoT (Wei et al., 2022) prompting was proposed to improve the ability of models to solve highly complex tasks – i.e., tasks that cannot be solved in a single step. The key idea is to introduce a chain of thoughts $t^{(1)}, \cdots, t^{(n)}$ as a meaningful and logical intermediate result that forms a bridge between $x$ and $y$, where $t^{(i)} \sim p_\theta\left(t^{(i)} | x, t^{(1)}, \cdots, t^{(i-1)}\right)$ and $t^{(i)}$ are sampled sequentially, and then the output $y$ is sampled from $y \sim p_\theta\left(y | x, t^{(1)}, \cdots t^{(n)}\right)$.

ToT (Yao et al., 2023) prompting was designed to help LLMs explore multiple reasoning paths over thoughts. Unlike CoT, it generates and manages a tree structure of intermediate reasoning steps known as thoughts. In its reasoning process, it allows the LLM to generate thoughts and evaluate each thought, then select which thoughts to reserve. This process is repeated until the LLM achieves the output goal $y$. $\mathrm{s} = \left(x, t^{(1)}, \cdots, t^{(i)}\right)$ denotes a reasoning path in the tree, which is essentially a partial solution reserved in the tree structure. The function $\mathrm{G}(p_\theta, s, k)$ denotes the thoughts generator, which samples $k$ thoughts from the distribution $t^{(i+1)} \sim p_\theta\left(t^{(i+1)} | s\right) = p_\theta\left(t^{(i+1)} | x, t^{(1)}, \cdots t^{(i)}\right)$. The function $\mathrm{V}(p_\theta, s^{i+1})$ then evaluates the state $s^{i+1} = \left(x, t^{(1)}, \cdots, t^{(i+1)}\right)$ and samples a value $v \sim p_\theta(v|s)$ for state $s^{i+1}$. The best thought is selected according to $t^{(i+1),j}$ by $v$, $j \in \{1,2, \cdots k\}$ where $k$ denotes the candidate thoughts sampled from $\mathrm{G}(p_\theta, s, k)$. ToT includes two search strategies, both of which traverse the decision tree. One is breadth-first-search; the other is depth-first-search.

As described in Chandrasekaran (1989) "A method can be a procedure where the sequencing of steps is all prespecified, but it can be more abstract; in Newell's problem space terminology (Newell, 1981), it can be a search in a problem space. In fact, such methods are the ones that are interesting from an AI point of view." Research on problem-solving with LLMs suggests that one of the main ways to solve a problem is through a tree-based search in problem space, where each node represents a partial solution (Yao et al., 2023). This tree-structured problem space is completely generated by the LLM through thought generation. However, this kind of thought generation is limited by the available training data. Plus, it carries cognition bias (Niu et al., n.d.). However, different problems can be modeled with different problem spaces, which directly impacts the difficulty of finding a solution (Simon, 1973; Simon & Newell, 1971). This insight reveals a key issue for problem-solving with LLMs: LLMs do not have the ability to adjust and select problem spaces. This can lead to either inefficient problem-solving or even failure to solve the problem at all. In other words, humans are still better at guiding the search for a solution through the problem space.

To address this shortcoming, I devised the Problem-Solving-Method-of-Thoughts (PSMoT). This is a paradigm that is incorporated into an LLM, which guides its reasoning path (Figure 1(d)). Unlike ToT, PSMoT relies on inherent guided reasoning frameworks (control flow) to engage in convergent reasoning or knowledge acquisition, ultimately arriving at a solution, rather than generating thoughts to search for solutions within the problem space. Further, the concept of thought in PSMoT and ToT differ. In PSMoT, thought plays the role of representing the input and output of the inference actions (see Figure 1(d)). The reasoning steps for this role are $r \sim p_\theta\left(r^{(i)} \mid x, r^{(1)}, \cdots, r^{(i-1)}, g^i\right)$, where $g^i$ is a predefined reasoning instruction or some piece of acquired knowledge for the thought $r^{(i)}$.

But why is PSMoT theoretically better than ToT? Consider ToT for a moment, where a thought is represented as $t^{(i+1)} \sim p_\theta\left(t^{(i+1)} \mid s\right)$. Here, the distribution of $p_\theta\left(t^{(i+1)} \mid s\right)$ has a possible thought set of $\mathrm{T} = \{t_1, t_2, \cdots, t_M\}$, which includes a feasible thought set of $T_f$, $\left|T_f\right| = \mathrm{m}$ and a non-feasible thought set of $T_{nf}$, $\left|T_{nf}\right| = M - m$. When the sum probability of feasible thoughts $P_f = \sum_{t_i \in T_f} \mathrm{p}(t_i)$ is very small, i.e., when $P_f \ll 1$, the probability that the non-repetitive thoughts sample $k$ will all be non-feasible is $P_{k,nf} = \left(1 - P_f\right)^k \approx 1 - kP_f$. In fact, this situation is common. Thought generation in an LLM generally follows a long-tailed distribution (Holtzman et al., 2020), which means only a few thoughts have a high probability of being sampled. And feasible thoughts are normally distributed across the long tail. Indeed, (Yao et al., 2023) show us that their models are still not capable of 100% accuracy in experiments with the 24 Game. This provides excellent evidence that it is quite common for some sampled thoughts to be infeasible solutions. By contrast, with PSMoT, solutions are found by relying on prior domain knowledge. In other words, domain knowledge can be used to specify the reasoning distribution for a given instruction $g^i$. This shapes the posterior distribution $p_\theta\left(r^{(i)} \mid x, r^{(1)}, \cdots, r^{(i-1)}, g^i\right)$ in a way that generates more feasible solutions than non-feasible solutions.

The following section shows how PSMoT is abstracted to heuristic classification.

### *3.2. Heuristic-Classification-of-Thoughts Prompting*

As a problem-solving method, heuristic classification in expert systems provides a structure through which to solve ill-structured problems, like diagnosis, catalog selection, and skeletal planning. Heuristic classification works by systematically associating data with pre-enumerated solutions through abstraction, heuristic association, and progressive refinement. Here, non-hierarchical conceptual relationships are prioritized as is representation-independent knowledge of the problem (Clancey, 1985; Fensel & Motta, 2001).

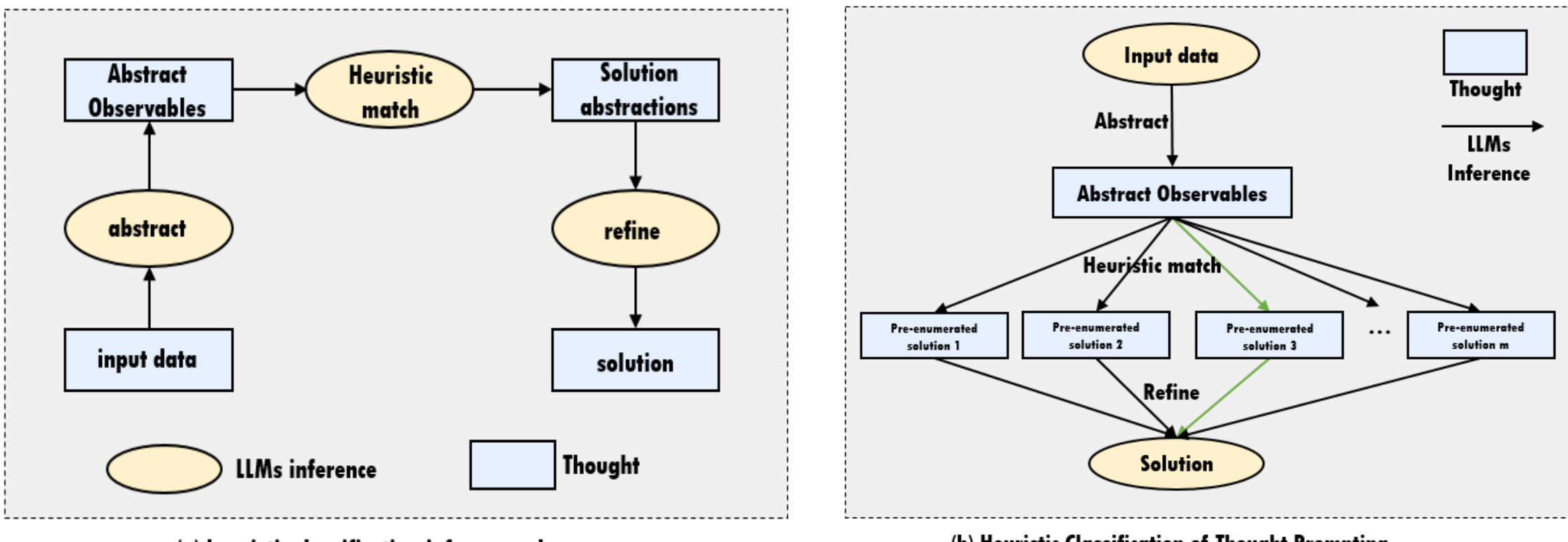

(a) heuristic classification inference scheme

(b) Heuristic Classification of Thought Prompting

Fig. 2. The heuristic classification inference scheme

As shown in Figure 2 (a), the inference structure of heuristic classification comprises three steps:

- Abstracting, which refers to extracting more general or abstract features or attributes from the input data so as to make those data suitable for subsequent reasoning. The aim is to better match pre-enumerated solutions; hence, inference relies on the association rules programmed for the pre-enumerated solutions.
- Heuristic matching, which refers to establishing direct, non-hierarchical associations between different hierarchies. The aim is to relate abstracted data to pre-enumerated solution categories through heuristics.
- Refining, which refers to refining the solution after matching it to a solution category. Based on the abstract solution, we get concrete solutions by learning more specific knowledge.

With heuristic classification, the problem space is modeled as several pre-enumerated solutions. Therefore, solving a problem involves abstracting the features from input data, then making a link between the existing solutions at either an abstract or the general level. The chosen solution is then refined to suit the current situation. To work effectively, heuristic classification must obviously normalize the underlying thinking and reasoning process so as to eliminate any cognitive bias on the part of the LLM.

This heuristic classification method is presented in PSMoT form in Figure 2(b). Here, an LLM is used to abstract the input data. Then the abstracted results are matched to candidate pre-enumerated solutions and further refined in the final step. In formal terms, three functions apply to the input data $x$, as follows.

- Abstract($x, p_\theta, g_{abstract}$): This instruction prompting $g_{abstract}$ on a sample returns an abstract or general description $a \sim p_\theta(r_{abstract} | x, g_{abstract})$, with which to better heuristically match a pre-enumerated solution.
- Matching ($a, x, p_\theta, g_{match}$): This function returns one or several pre-enumerated solutions as candidate abstract solutions for further refinement. The matching is a sample process $c \sim p_\theta(r_{class} | a, x, g_{match})$ ,where $g_{match}$ should include the acquired pre-enumerated solutions matched to the instruction prompt, and $a$ is the abstract description from Abstract(), which returns the matched result $c$.
- Refinement($c, x, p_\theta, g_{refine}$): This function generates final solutions by referring to the abstract solutions. The LLM samples the solutions $s \sim p_\theta\left(r_{solution} \middle| c, x, g_{refine}\right)$, where $g_{refine}$ is the refinement prompt and $c$ is the abstract solution derived from Matching().

Algorithm 1 below sets out how these three functions combine to complete the HCoT prompting process illustrated in Figure 2(b).

| Algorithm 1 HCoT($x, g_{abstract}, g_{match}, g_{refine}$) |
|---|
| Require: input $x$, LLM generation distribution $p_\theta$, through Abstract(), Matching(), Refinement() function generate solution<br>Description = Abstract($x$, $p_\theta$, $g_{abstract}$)<br>Abstract-solutions = Matching(Description, $x$, $p_\theta$, $g_{match}$)<br>Solution = Refinement(Abstract-solutions, $x$, $p_\theta$, $g_{refine}$)<br>Return Solution |

The next section details the experiments, in which we test HCoT on two inductive reasoning tasks and the 24 Game.

# 4. Experiments

## *4.1. Experiments with the 24 Game*

**The problem**: The 24 Game (Yao et al., 2023) is a popular mathematical card game. The objective is simple: using four given numbers and the four basic operations (addition, subtraction, multiplication, and division), combine them to form the number 24. Each number must be used exactly once, but the operations, including parentheses, can be applied in any order to dictate the sequence. I compared HCoT's performance to 1362 games collected from 4nums.com (*Classical Math Game, Use All 4 Numbers and + - × / to Make 24!*, n.d.). Unlike Yao et al. (2023), I used all 1362 games for testing instead of just 100.

The 24 Game is a well-defined problem (Russell & Norvig, 2016) because the initial state, the operations, the goal, and the costs can be easily specified. Additionally, ToT approaches have proven to be good at mastering this game. Therefore, it makes sense to compare HCoT to ToT-BFS to assess performance. I believe HCoT will have an advantage in terms of time costs and budget. In other words, HCoT will represent a token-efficient and budget-aware alternative for this deliberate reasoning game.

**The pre-enumerate solution (problem space)**: Searching the website, I found the six most common solutions to the 24 Game(*General Methods and Examples for Solving the 24 Game*, n.d.). These appear in Table 1.

Table 1 Common patterns frequently used in the 24 Game

| | Solution patterns and examples for the 24 Game |
|---|---|
| 1 | (a - b) * (c + d)<br>Example: (10 - 4) * (2 + 2) = 24 |
| 2 | (a + b) / c * d<br>Example: (10 + 2) / 2 ×4 = 24 |
| 3 | (a - b / c) * d<br>Example: (3 - 2 / 2) * 12 = 24 |
| 4 | (a + b - c) * d<br>Example: (9 + 5 - 2) * 2 = 24 |
| 5 | a * b + c - d<br>Example: 11 * 3 + 1 - 10 = 24 |
| 6 | (a - b) * c + d |

| | Example: (4 - 1) * 6 + 6 = 24 |
|---|---|

In addition to these six patterns, another 10 patterns were included in the model as a supplement to test the impact of expanding the solution space.

**Experimental Design**: The heuristic classification pipeline was organized in two stages.

- Abstraction: each instance of the 24 Game (i.e., its four numbers) was first converted into an abstract representation.
- Matching & refinement: a description of the abstraction was then compared with a catalog of pre-enumerated solution templates (see Table 6). The best-matching template was selected and refined to yield a concrete solution.

**Baselines:** HCoT was compared to standard input-output (IO) prompting and CoT prompting (see Yao et al., 2023), each of which had five in-context examples. The ToT-prompting was implemented as a classic ToT prompt, being: "Imagine three different experts are answering this question. All experts will write down…"(*Tree of Thoughts (ToT)*, n.d.). ToT-BFS was also tested with a beam width of 5. This means the LLM used a tree-based search strategy and retained the top five most promising states at each step. All experiments involved GPT-4o-2024-08-06 as the base model, with a temperature of 1.0. So as to test the limits of ToT-BFS on all 1362 instances of the 24 Game, an additional experiment was conducted with the temperature set to 0.7.

**HCoT setup:** I conducted five experiments, each with a distinct configuration of the HCoT framework:

- HCoT (6-patterns): This only uses the six pre-enumerated solution patterns listed in Table 1, all presented within a single prompt.
- HCoT (16-patterns): This expands the solution set to 16 pre-enumerated patterns (6 from Table 1 + the other 10), with all patterns provided in one prompt.
- HCoT (16-patterns-Split-6+10): Splits the 16 patterns across two sequential prompts – the first containing the 6 patterns from Table 1, and the second containing the remaining 10. The two prompts are executed in sequence; if a match is found in the first prompt, the second is skipped.
- HCoT (16-patterns-best-of-2): Samples the refinement step twice using the full set of 16 patterns and stops as soon as a valid solution is found.
- HCoT (16-patterns-best-of-4): Similar to best-of-2, but samples up to four times in the refinement step, stopping at the first successful match.

Table 2 the 24 Game experiments result

| Approach | Accuracy | Generate token | Prompt token |
|---|---|---|---|
| IO-Prompt | 77/1362=5.65% | 65764 | 709592 |
| CoT Prompt | 130/1362=9.54% | 385411 | 709602 |
| ToT-Prompt | 43/1362=3.16% | 176947 | 256056 |
| ToT-BFS (b=5, temperature=0.7) | 758/1362=**55.65%** | 7283699 | 27517380 |
| ToT-BFS (b=5, temperature=1) | 619/1362=45.45% | 5806078 | 24233257 |
| HCoT(6-patterns) | 443/1362=32.53% | 1750725 | 3235082 |
| HCoT(16-patterns) | 331/1362=24.3% | 1952310 | 5334341 |
| HCoT(16-patterns-Split-6+10) | 497/1362=36.49% | 2880077 | 6931512 |
| HCoT(16-patterns-best-of-2) | 372/1362=27.31% | 2877436 | 8526257 |
| HCoT(16-patterns-best-of-4) | 500/1362=36.71% | 4460751 | 14064991 |

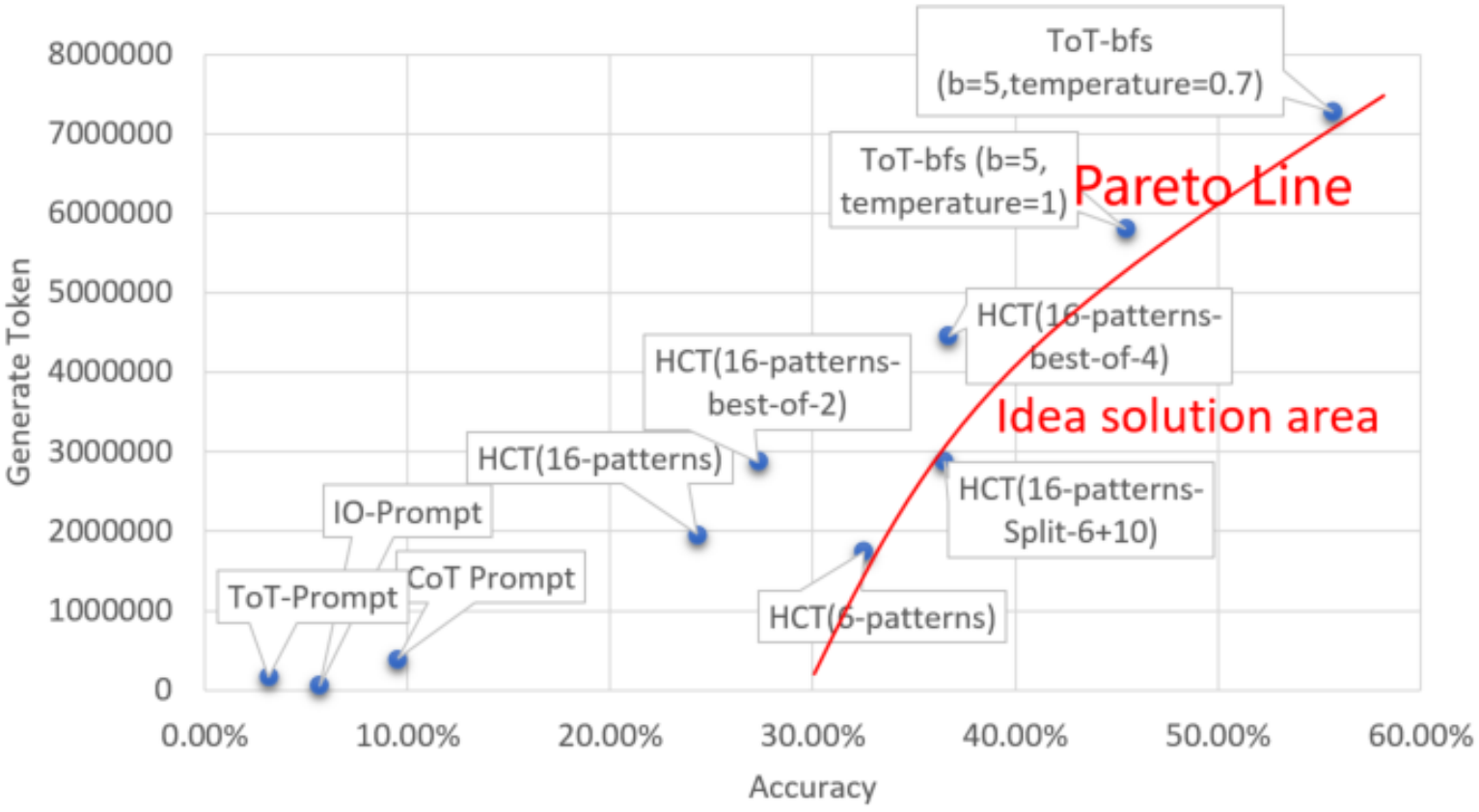

Fig. 3. Scatter plot of accuracy versus number of generated tokens across the nine experiments.

**Results**: The experiments demonstrate the superiority of the HCoT framework in balancing accuracy and token efficiency on the Pareto frontier. Notably, HCoT (16-patterns-Split-6+10) achieves 36.49% accuracy with only 2.88M tokens, outperforming ToT-BFS in efficiency-critical scenarios with 55.65% accuracy but 7.28M tokens. HCoT's structured pattern-guided search avoids the computational overhead of ToT-BFS's brute-force exploration. Expansion strategies reveal critical trade-offs: HCoT (16-patterns) underperforms 6-patterns (24.3% vs. 32.53%), indicating noise from excessive patterns, while 16-patterns-separate-2 succeeds by partitioning patterns into prioritized stages, filtering noise early. Further, best-of-2 only achieves accuracy 27.31% while best-of-4 marginally improves accuracy (36.71%) at high token costs, highlighting the improvement is from the partitioning strategy, not from more sampling. ToT-BFS's sensitivity to temperature is evident: lowering the temperature to 0.7 boosts accuracy from 45.45% to 55.65% but increases token usage by 25.5%, emphasizing its reliance on deterministic paths for well-defined problems. In conclusion, HCoT's hybrid of pattern-driven reasoning and adaptive strategies achieves competitive accuracy with superior efficiency, positioning it as a scalable solution for structured reasoning tasks.

*4.2. Experiments on 1D-ARC*

**The problem**: The 1D-ARC dataset, introduced by Xu et al. (2023), is a one-dimensional, LLM-oriented adaptation of the Abstraction and Reasoning Corpus (ARC). Whereas the original ARC poses abstract reasoning puzzles on 2-D grids, 1D-ARC reformulates them as transformations over linear arrays—a representation that LLMs handle more naturally than pixelated images. Like its parent corpus, 1D-ARC measures a model's ability to infer and generalize rules from only a few examples, but it does so in a format that better matches the sequence-processing strengths of LLMs. The collection spans 18 transformation categories and 901 individual tasks. Every task supplies three input-output training pairs plus one hold-out test pair, challenging the model to extrapolate the underlying rule and produce the correct test output after seeing only the three demonstrations. A key advantage of this dataset is that every example is explicitly annotated with its corresponding abstract solution. This problem is not a well-defined problem. These experiments involved the base model GPT-4o-2024-08-06 and O3-mini-2025-01-31 with temperatures of 0.7 and 1.0.

**The pre-enumerated solutions (problem space):** Table 3 shows 2 of the 18 pre-enumerated solutions solutions alongside an example of each. These 18 solutions (no repeats) cover all the tests.

Table 3 1D-ARC pre-enumerate solution examples

| | **Pre-enumerated solution** |
|---|---|
| 1 | Task 1: 1d_denoising_1c<br>Rule: Retain the longest contiguous segment of the same color and remove all other isolated or scattered colored pixels (considered noise).<br>- Only the primary color block is preserved.<br>- All other colored elements are eliminated.<br>Example:<br>{'input': [[0, 0, 5, 0, 0, 0, 0, 0, 5, 5, 5, 5, 5, 5, 5, 5, 5, 5, 0, 0, 0, 0, 5, 0, 0, 0]],<br>'output': [[0, 0, 0, 0, 0, 0, 0, 0, 5, 5, 5, 5, 5, 5, 5, 5, 5, 5, 0, 0, 0, 0, 0, 0, 0, 0]]} |
| 2 | Task 2: 1d_denoising_mc<br>Rule: Recolor isolated pixels within the color segments to match neighboring values, effectively blending them into the block.<br>- Retain full segments; integrate anomalies.<br>Example:<br>{'input': [[5, 5, 5, 6, 5, 5, 5]],<br>'output': [[5, 5, 5, 5, 5, 5, 5]]} |
| … | … |

**Experimental design:** The induction pipeline is divided into three stages: description, matching, and refinement.

- Description: For every training input-output pair, extract a compact set of features that capture its essential transformation; these features serve as a heuristic signature.
- Matching: Using this signature, search a library of pre-enumerated solution programs and select the candidate that best fits the training examples.
- Refinement: The selected program is then applied, without further tuning, to the test input to generate the predicted test output.

**Baselines:** I compared HCoT to similar baselines as those used in the 24 Game experiments. However, since this task is not well-defined, ToT-BFS was excluded from the evaluation, and Auto CoT (zero-shot) (Z. Zhang et al., 2023) was used instead of CoT. The experiments were conducted using two base models – one optimized for efficiency (GPT-4o) and the other for strong reasoning (GPT-o3-mini). This enabled me to assess performance across both lightweight and advanced reasoning settings.

**HCoT setup:** Two experiments were conducted, each using a different base model, with each implemented as a three-step process guided by prompts. Thanks to the availability of labeled data, I was able to evaluate the performance of the heuristic matching and refinement processes separately. Matching accuracy was defined as the proportion of cases where the system correctly identified the intended pre-enumerated solution. Match-and-solve accuracy refers to the proportion of correctly matched cases that also produced the correct test output. In other words, it measures the success rate of refinement conditioned on successful matching – similar in spirit to Precision or an F1-score.

Table 4 1D-ARC experiments result

| Base model | Approach | Temperature | Accuracy | Matching accuracy | Match-and-solve accuracy |
|---|---|---|---|---|---|
| gpt-4o- | IO-Prompt | 0.7 | 397/901=44.06% | None | None |

| 2024-08-06 | Auto CoT-Prompt | | 346/901=38.40% | None | None |
|---|---|---|---|---|---|
| | ToT-Prompt | | 281/901=31.19% | None | None |
| | HCoT | | 429/901=**47.61%** | 524/901=58.16% | 367/524=70.04% |
| | IO-Prompt | 1 | 361/901=40.06% | None | None |
| | Auto CoT-Prompt | | 361/901=40.06% | None | None |
| | ToT-Prompt | | 251/901=27.86% | None | None |
| | HCoT | | 393/901=**43.62%** | 545/901=60.49% | 333/545=61.1% |
| o3-mini-2025-01-31 | IO-Prompt | 1 | 747/901=82.91% | None | None |
| | ToT-prompt | | 774/901=85.90% | None | None |
| | HCoT | | 788/901=**87.46%** | 745/901=82.69% | 677/745=90.87% |
| | IO-Prompt | 0.7 | 775/901=**86.02%** | None | None |
| | ToT-prompt | | 773/901=85.79% | None | None |
| | HCoT | | 770/901=85.46% | 747/901=82.91% | 676/747=90.50% |

**Results:** The HCoT framework outperformed all the traditional methods (IO-Prompt, Auto CoT-Prompt, ToT-Prompt) in the1D-ARC experiments, achieving 47.61% accuracy with GPT-4o and 87.46% with O3-mini. Notably, HCoT excels in solution matching (60.49% to 82.91%) and refinement (70.04% to 90.87% match-and-solve success), demonstrating robustness across the models. With GPT-4o, HCoT improved stability in long-horizon reasoning by structuring the problem-solving, while with O3-mini, it efficiently optimized the solution execution without sacrificing accuracy, underscoring its adaptability to both resource-intensive and lightweight settings. In addition, the analysis reveals a critical bottleneck in the matching stage, as evidenced by the disparity between the matching accuracy (58.16% for GPT-4o; 82.91% for O3-mini) and the match-and-solve accuracy (70.04% for GPT-4o; 90.87% for O3-mini). While the refinement stage demonstrates high success rates conditioned on correct matching (e.g., 90.50% for O3-mini), the relatively lower matching accuracy indicates that the system frequently fails to identify the correct pre-enumerated solution. This gap underscores that improvements in heuristic-guided pre-enumerate solution matching would yield disproportionate gains in overall performance, as downstream refinement is already robust. This bottleneck is exacerbated in weaker models (e.g., GPT-4o), where matching failures directly limit the applicability of even effective refinement logic. Notably, while a higher temperature (1.0) benefited HCoT with GPT-4o (47.61% → 43.62% accuracy decline), it inversely harmed performance with O3-mini (HCoT: 85.46% at 0.7 vs. 87.46% at 1.0). This discrepancy suggests temperature sensitivity varies by model architecture: GPT-4o's refined reasoning may suffer from over-diversification under high temperatures, disrupting structured HCoT workflows, whereas o3-mini leverages controlled stochasticity to enhance heuristic exploration. However, elevated temperatures degrade other methods (e.g., IO-Prompt dropped to 40.06% on GPT-4o at 1.0), indicating HCoT's unique capacity to harness temperature-driven diversity without sacrificing coherence. Crucially, HCoT maintained superiority even on advanced models like O3-mini (87.46% vs. baseline 86.02%), highlighting its scalable efficacy in balancing exploration and precision across reasoning paradigms.

*4.3. Experiments on List Function*

**The problem:** The list function dataset was first introduced in Rule (2020). Since then, it has been used to evaluate the inductive reasoning capabilities of LLMs several times (Qiu et al., 2024; R. Wang et al., 2024). The dataset comprises 250 tasks, each containing 8 training examples and 8 test instances. The training examples consist of input-output array pairs governed by an underlying transformation rule. The test set follows the same format but includes both inputs and their corresponding labeled outputs, which serve as ground truths for evaluation. This task is not well-defined in the traditional AI sense: it lacks an explicit definition of states and operations, making it impossible for algorithms like ToT-BFS to construct a formal search space. Instead, models must perform an inductive program synthesis directly from the examples without a predefined symbolic structure. I used three base models in these experiments: GPT-4o-2024-08-06, DeepSeek-V3 (DeepSeek-AI et al., 2024), and DeepSeek-R1 (DeepSeek-AI, 2024), all with a temperature of 0.8.

**The pre-enumerated solutions (problem space):** The first step was to manually define a set of 16 abstract solution templates derived from the list function tasks. These templates cover a subset of the 250 tasks, while the remaining tasks are handled by the LLMs without access to predefined solutions. In addition, I introduced 8 supplementary solution templates for the comparative experiments, for greater generalization beyond the initial set. Table 5 shows an excerpt of the pre-enumerated solution templates.

Table 5 example of the pre-enumerated solutions for the List Function dataset

| | The pre-enumerated solutions |
|---|---|
| 1 | FixedIndexSelector – return element at fixed index m (empty if length ≤ m). |
| 2 | SliceExtractor – return slice [start:end:step] (empty if length ≤ index ≤ end). |
| 3 | ExtremumPicker – return max or min of the sequence. |
| 4 | FixedIndexSummer – sum elements at specific indices. |
| 5 | FixedIndexMultiplier – multiply elements at specific indices. |
| | … |

**HCoT Experiment design:** Again, the induction pipeline was organized into three steps: description, matching, and refinement.

- Description. Given 8 training input-output pairs, the LLM is prompted to observe and abstract structural relationships between inputs and outputs, generating a high-level description of the transformation.
- Matching. Based on the generated abstract description, the LLM selects the most appropriate candidate from the set of pre-enumerated solutions, assigning concrete parameter values to instantiate the scheme. If no candidate aligns with the observed pattern, it returns a fallback response indicating that no abstract scheme fits.
- Refinement. The selected abstract scheme is verified and applied to each test input using the inferred parameters, which generates a corresponding output. If no matching abstract scheme is found, the LLM is instructed to, first, generate a concise, task-specific transformation rule that explains the training behavior and, then, apply this custom rule to the test inputs.

**Baselines:** HCoT was compared to the same set of baselines as used in the 1D-ARC experiments. To evaluate performance across different model types, experiments were conducted with three base models: GPT-4o and DeepSeek-V3, representing general-purpose language models, and the third model DeepSeek-R1, which was chosen for its strong reasoning capabilities.

**HCoT setup:** For this experiment, I created five variations of the HCoT framework so as to evaluate performance across different base models:

- 16-solutions: Uses 16 pre-enumerated solutions to assess baseline performance.
- 24-solutions: Uses 24 pre-enumerated solutions to test whether increased abstraction leads to improved performance.
- Split-12+12: Splits the 24 solutions into two sets of 12, requiring the model to perform matching twice. This setup tests whether fewer candidate solutions per step improve matching accuracy.

- Split-16+8: Separates the 24 solutions into groups of 16 and 8. This is used to compare with the 12+12 split to examine whether different partitioning schemes affect performance.
- 24-solutions-best-of-2: Samples from the full set of 24 solutions twice and selects the better match. This is used to ensure that the performance gains from the split setups are not simply due to the model sampling twice.

Table 6 the List Function experiments result

| Base-model | Problem-solving approach | Accuracy |
|---|---|---|
| GPT-4o-2024-08-06 | IO-Prompt | 87/250=34.8% |
| | Auto CoT-Prompt | 92/250=36.8% |
| | ToT-Prompt | 87/250=34.8% |
| | HCoT(16-solutions) | 106/250=42.4% |
| | HCoT(24-solutions) | 98/250=39.2% |
| | HCoT(24-solutions-best-of-2) | 98/250=39.2% |
| | HCoT(Split-12+12) | 102/250=40.8% |
| | HCoT(Split-16+8) | 108/250=**43.2%** |
| DeepSeek V3 | IO-Prompt | 151/250=60.4% |
| | ToT-Prompt | 161/250=64.4% |
| | Auto CoT-Prompt | 151/250=60.4% |
| | HCoT(16-solutions) | 163/250=65.2% |
| | HCoT(24-solutions) | 165/250=66.0% |
| | HCoT(Split-12+12) | 178/250=**71.2%** |
| | HCoT(24-solutions-best-of-2) | 166/250=66.4% |
| | HCoT(Split-16+8) | 173/250=69.2% |
| DeepSeek R1 | IO-Prompt | 190/250=76.0% |
| | Auto CoT-Prompt | 183/250=73.2% |
| | ToT-Prompt | 136/250=54.4% |
| | HCoT(16-solutions) | 194/250=77.6% |
| | HCoT(24-solutions) | 198/250=79.2% |
| | HCoT(Split-12+12) | 195/250=78.0% |
| | HCoT(24-solutions-best-of-2) | 197/250=78.8% |
| | HCoT(Split-16+8) | 203/250=**81.2%** |

**Result:** As shown in Table 6, the results demonstrate that the HCoT framework significantly enhances task-solving capabilities, particularly excelling when applied to models with strong structured abstraction abilities, such as DeepSeek-R1. Under the HCoT configuration (Split-16+8), DeepSeek-R1 achieved an accuracy of 81.2%, representing a 5.2 percentage point improvement over the traditional IO-Prompt. Moreover, accuracy steadily improved with the pre-enumerated solution expansion (from 77.6% at 16 solutions to 79.2% at 24 solutions). Most notably, the partitioning strategy had a substantial impact on performance: DeepSeek-V3 reached 71.2% accuracy under a Split-12+12 configuration (10.8% improvement over IO-Prompt), while R1 performed best with a 16+8 split (81.2%). In order to validate that the improvements were due to the partitioning strategy, I conducted more experiments, comparing the 24-solutions-best-of-2 configuration with the 24 pre-enumerate solutions strategy. The results leave no doubt that the expansion solution outperforms the baseline; however, partitioning strategies achieve superior performance. By contrast, GPT-4o performed relatively poorly, with a maximum accuracy of only 43.2%, highlighting its limited capacity for abstract solution matching and refining. Overall, HCoT outperformed the traditional methods like Auto-CoT and ToT, benefiting from its stepwise abstraction and verification mechanisms. For example, R1 achieved a 26.8% improvement over ToT-Prompt, validating the effectiveness of a structured inductive pipeline.

## 5. Discussion

These results demonstrate that HCoT can enhance a model's problem-solving abilities with both well-structured and ill-structured problems. This improvement may be attributed to HCoT's approach of integrating prior knowledge to structure the problem space. By doing so, HCoT systematically organizes the problem space and controls the reasoning process to converge on pre-enumerated solutions. This structured approach helps mitigate reasoning biases, thereby contributing to more effective problem-solving.

The key step in HCoT is heuristic matching. This approach transforms solution generation into solution selection. In practice, X. Zhang et al. (2023) report that LLMs perform better with objective questions than subjective ones, which means LLMs perform better at multiple-choice questions than open-ended queries. In essence, ToT formulates problems as an open-ended question, while the key to HCoT is that it formulates problems as a multiple-choice question through heuristic matching. One notable limitation of HCoT lies in its reliance on pre-enumerated solutions, which can inadvertently steer the search process towards the local optima so as to avoid an exhaustive search (Buchanan & Smith, 1988). Clancey (1985) refers to this as an opportunistic search. By structuring the problem space around these predefined solutions, HCoT might prioritize similar or closely matching options during the solution search. This inclination towards a local search can stop the model from exploring broader solution spaces, potentially overlooking more innovative or globally optimal solutions. Consequently, while HCoT's approach offers efficiency and structure, it is essential to be mindful of this limitation to ensure a comprehensive and balanced search for optimal solutions. This limitation is also explained by the experiment results which show that expanding the set of pre-enumerated solutions can, in some cases, improve success rates. These findings underscore the potential for fostering collaborative problem-solving endeavors between human experts and LLMs. They especially promote applying alternative problem-solving methods when using LLMs.

The other significant finding is that using partitioning strategies involving pre-enumerated solutions can improve performance. For example, with the two induction reasoning tasks, expanding the solution set increased the model's success rate. Yet, that said, with the 24 Game in Table 2, it paradoxically reduces computational efficiency and decreases the success rate in practice. However, when separating the pre-enumerated solutions from one prompt to two, success rates increased. This result aligns with findings from the literature on multi-classification problems: as the number of classes increases, the average accuracy decreases (AlZoman & Alenazi, 2021; Van Thinh et al., 2019). Thus, the heuristic matching step can be thought of as a multi-classification problem, where separating the pre-enumerated solutions decreases the number of classes associated with each prompt, in turn, increasing the accuracy of matching. Note, however, that partitioning strategies are limited by the token costs, where increases in the number of prompts increase the token generation.

Overall, these experiments indicate that integrating hierarchical or coarse-to-fine reasoning approaches into the problem-solving process of LLMs can significantly enhance success rates.

## 6. Conclusion

In this study, we systematically explored the synergy between LLMs and structured methodologies to streamline reasoning processes, enforce algorithmic control, and drive convergence toward focused problem-solving pathways, thereby deriving actionable insights into the cognitive dynamics of ill-structured problem-solving. Our findings demonstrate that Heuristic-Classification-of-Thoughts (HCoT) prompting significantly enhances the problem-solving efficacy of LLMs across both general-purpose base models (e.g., DeepSeek-V3, GPT-4o) and specialized long-chain reasoning architectures (e.g., GPT-o3, DeepSeek-R1). Notably, pre-enumerated solution templates, as a form of prior knowledge injection, critically enhance the models' capacity to structure abstract problem spaces and guide solution exploration.

While this research confirms that structuring problem spaces through abstract solution templates accelerates search efficiency, it is clear that even when the problem space is well-structured and comprehensive, using long-chain architectures still fails to perfectly resolve all issues. This highlights a persistent gap between LLMs and human reasoning capabilities. It underscores the need for humans to leverage methods rooted in prior knowledge to further narrow this gap. This tension between model capabilities and human-like reasoning emphasizes the importance of dynamically balancing pre-existing domain knowledge with the context-aware knowledge generation capabilities inherent in LLMs. By reframing prior knowledge as a scaffolding mechanism that supports reasoning, rather than rigidly constraining it, this framework mitigates reasoning biases stemming from training data imbalances, while preserving the adaptability and generative potential of AI.

Ultimately, this work advances the frontier of LLM-driven problem-solving for ill-defined challenges, offering a methodological blueprint for prompt engineering that strategically leverages prior knowledge to reduce cognitive biases in inner reasoning workflows. we anticipate these insights will catalyze broader interdisciplinary explorations into the interplay between human-guided heuristics and machine-generated intuition.

## Appendix A

Table A.1 Instruction prompt for Matching & refinement step in HCoT (16-patterns-Split-6+10) of the 24 Game

| Prompt_matching_6_abstract_solution=""" Match and Try the following 7 patterns: |
|---|

pattern 1: (a - b) * (c + d)
Example: (10 - 4) * (2 + 2) = 24

pattern 2: (a + b) / c * d
Example: (10 + 2) / 2 ×4 = 24

pattern 3: (a - b / c) * d
Example: (3 - 2 / 2) * 12 = 24

pattern 4: (a + b - c) * d
Example: (9 + 5 - 2) * 2 = 24

pattern 5: a * b + c - d
Example: 11 * 3 + 1 - 10 = 24

pattern 6: (a - b) * c + d
Example: (4 - 1) * 6 + 6 = 24

pattern 7: No solution matches
if no solution worked, try to solve it by yourself.
"""

Prompt_matching_10_abstract_solution=""" Match and Try the following 11 patterns:

pattern 1: c = d , a * c - b * d = (a - b) * c
Example: 9 * 4 - 3 * 4 = (9 - 3) * 4 = 24

pattern 2: c = d ,a * c + b * c = (a + b) * c
Example: 1 * 4 + 5 * 4 = (1 + 5) * 4 = 24

pattern 3: a + b + c + d = 24
Example: 1 + 2 + 9 + 12 = 24

pattern 4: a * b * c * d = 24
Example: 1 * 2 * 3 * 4 = 24

pattern 5: use the four numbers to construct 3 and 8 through addition, subtraction, multiplication, or division, and then multiply them to get 24.
Example: (4 - 1) *(1 * 8) = 24

pattern 6: use the four numbers to construct 2 and 12 through addition, subtraction, multiplication, or division, and then multiply them to get 24.
Example: 1+1=2, 2*6 =12 -> (1+1)*2*6=24

pattern 7: use the four numbers to construct 4 and 6 through addition, subtraction, multiplication, or division, and then multiply them to get 24.
Example: (8/2) = 4 and (2 + 4) = 6, (8/2)* (2+4)=24

pattern 8: use the four numbers to construct 1 and 24 through addition, subtraction, multiplication, or division, and then multiply them to get 24.
Example: (9/9) = 1 and 12 + 12 = 24, (9/9)*(12+12)=24

pattern 9: use the four numbers to construct pairs of numbers that sum or subtract to 24, such as 21 + 3, 20 + 4, 18 + 6, 16 + 8, 12 + 12, 27 − 3, 28 − 4, or 30 − 6.
Example: (1+2)=3 and 3*7=21 -> 1+2+3*7=24

pattern 10: a * (c + d / b) = 24 or a * (c - d / b) =24
Example:(3 +  3 / 7) * 7 = 24
Example:4 / (1 - 5 / 6) = 24

pattern 11: No solution matches
if no solution worked, try to solve it by yourself.
"""

Table A.2 Instruction prompt for Matching step in HCoT(Split-16+8) of List Function

Prompt_Matching_1="""
Using the abstract description you just produced, select possible schemes (with precise parameter values) from the catalogue below that best fits the observed pattern, and output only the scheme name plus its parameters. if no scheme matches returns no abstract scheme fits.
1) FixedIndexSelector – return element at fixed index m (empty if length ⩽ m).
2) SliceExtractor – return slice [start:end:step].(empty if length ⩽ index ⩽ end)
3) ExtremumPicker – return max or min of the sequence.
4) FixedIndexSummer – sum elements at specific indices.
5) FixedIndexMultiplier – multiply elements at specific indices.
6) SimpleSwap – swap two specified indices.
7) ValueBasedSwap – swap two fixed indices only if a size condition holds.
8) FixedIndexRemover – remove element at fixed index m.
9) ValueChanger – replace the values at selected indices with new values.
10) SliceReverser – reverse the order of a specified slice in place.
11) ThresholdFilter – keep or discard elements greater/less than threshold T.
12) ScalarArithmetic – add, subtract, multiply, or divide each element by constant k.
13) DuplicateFilter – remove repeated values while preserving their first occurrence.
14) SliceSumInserter – insert the sum of a slice at fixed position p.
15) SliceRemover – delete a slice of length L starting at position pos.
16) EdgeDuplicateTrimmer – if the first or last two elements are identical, delete that pair.

Return — Use the exact template:
******
1) scheme_1
2) scheme_2
...
******

for example
******
1) FixedIndexSelector

8) FixedIndexRemover
******
"""

Prompt_Matching_2="""
Using the abstract description you just produced, select possible schemes (with precise parameter values) from the catalogue below that best fits the observed pattern, and output only the scheme name plus its parameters. if no schem matches returns no abstract scheme fits.

17) HeadTailChooser – delete a head or tail segment of length length_to_drop based on endpoint sizes.
18) SliceSumReinserter – compute the sum of slice [start : end] and re-insert it at position p.
19) FixedSliceRemover – remove a slice of length L starting at position pos (alias of SliceRemover).
20) TwinEdgeRemover – if the first or last two adjacent numbers are equal, remove that pair.
21) AdaptiveEdgeSliceRemover – drop a head or tail slice of length L when the specified criterion holds.
22) RelativeValueSwap – select two elements by a size-based rule and swap them.
23) SafeInserter – insert value at position p; if p is out of bounds, append instead.
24) LengthReporter – return the sequence length (as an integer or singleton list).

Return — Use the exact template:
******
1) scheme_1
2) scheme_2
...
******

for example
******
13) DuplicateFilter
16) EdgeDuplicateTrimmer
******
"""

Table A.3 Instruction prompt for Matching step in HCoT of 1D-ARC

Prompt_matching = """
Match the transformation rule between the input and output in the training data to one of the following 18 types:

### Task 1: 1d_denoising_1c
Rule: Retain the longest contiguous segment of the same color and remove all other isolated or scattered colored pixels (considered noise).
- Only the primary color block is preserved.
- All other colored elements are eliminated.

Example:
{'input': [[0, 0, 5, 0, 0, 0, 0, 0, 5, 5, 5, 5, 5, 5, 5, 5, 5, 5, 0, 0, 0, 0, 5, 0, 0, 0]],
'output': [[0, 0, 0, 0, 0, 0, 0, 0, 5, 5, 5, 5, 5, 5, 5, 5, 5, 5, 0, 0, 0, 0, 0, 0, 0, 0]]}

### Task 2: 1d_denoising_mc

Rule: Recolor isolated pixels within color segments to match neighboring values, effectively blending them into the block.
- Retain full segments; integrate anomalies.

Example:
{'input': [[5, 5, 5, 6, 5, 5, 5]],
 'output': [[5, 5, 5, 5, 5, 5, 5]]}

### Task 3: 1d_fill
Rule: Fill gaps between two color segments using the color of adjacent segments.
- Produces continuous blocks from previously disconnected segments.
Example:
{'input': [[0, 0, 2, 0, 0, 2, 0]],
 'output': [[0, 0, 2, 2, 2, 2, 0]]}

### Task 4: 1d_flip
Rule: This task involves reversing the order of a single contiguous color block within a 1D array. The following constraints must be respected:
- Only one contiguous block of non-zero color values is present in the array.
- The block must retain its exact colors and length, but their positions must be reversed.
- The start and end borders (zeros outside the color block) must remain unchanged.
- The overall length of the array remains the same.
This results in a horizontal mirror flip of the colored segment, while preserving all surrounding structure.
Example:
{'input':  [[ 0, 0, 0, 0, 0, 3, 5, 5, 5, 5, 5, 5, 5, 5, 5, 5, 5, 5, 0, 0, 0, 0, 0, 0, 0]],
 'output': [[ 0, 0, 0, 0, 0, 5, 5, 5, 5, 5, 5, 5, 5, 5, 5, 5, 5, 3, 0, 0, 0, 0, 0, 0, 0]]}
{'input':  [[ 0, 0, 0, 2, 3, 3, 3, 3, 3, 3, 3, 3, 3, 3, 3, 3, 3, 3, 0, 0, 0, 0]],
 'output': [[ 0, 0, 0, 3, 3, 3, 3, 3, 3, 3, 3, 3, 3, 3, 3, 3, 3, 2, 0, 0, 0, 0]]}

### Task 5: 1d_hollow
Rule: Preserve only the borders of color segments, replacing the internal pixels with background.
- Start and end of each block are retained.
Example:
{'input': [[0, 3, 3, 3, 0]],
 'output': [[0, 3, 0, 3, 0]]}

### Task 6: 1d_mirror
Rule: This task involves performing a horizontal mirroring of the first contiguous color block with respect to the second color block in the array. The mirrored version of the first block is relocated to the opposite side of the second block, while all other positions are filled with zeros.
The following constraints must be respected:
- The array contains exactly two non-zero segments.
- The first segment is the one being mirrored and moved.
- The second segment remains fixed in position.
- The mirrored segment is inserted at an equal offset on the other side of the second segment (as if reflected across it).
- The overall length of the array must remain unchanged.

This creates a visual "mirror" effect across the second segment.
Example:
{'input':  [[6, 6, 6, 6, 0, 4, 0, 0, 0, 0, 0, 0, 0, 0, 0]],
 'output': [[0, 0, 0, 0, 0, 4, 0, 6, 6, 6, 6, 0, 0, 0, 0]]}
{'input':  [[0, 0, 8, 8, 8, 8, 0, 4, 0, 0, 0, 0, 0, 0, 0]],
 'output': [[0, 0, 0, 0, 0, 0, 0, 4, 0, 8, 8, 8, 8, 0, 0]]}

### Task 7: 1d_move_1p
Rule: This task involves a **single contiguous color block** moving **exactly one position to the right** within the array. The following constraints must be respected:
The following constraints must be respected:
- Only **one** color block is present in the array.
- The block must retain its original **color** and **length**.
- The **overall length of the array must remain unchanged**.

Example valid transformations:
{'input': [[4, 4, 4, 4, 4, 4, 4, 4, 4, 4, 4, 4, 4, 4, 4, 4, 4, 4, 4, 4, 4, 4, 4, 0, 0, 0, 0]],
 'output': [[0, 4, 4, 4, 4, 4, 4, 4, 4, 4, 4, 4, 4, 4, 4, 4, 4, 4, 4, 4, 4, 4, 4, 4, 0, 0, 0]]}
{'input': [[0, 0, 0, 0, 0, 0, 0, 0, 0, 0, 0, 0, 0, 0, 2, 2, 2, 2, 0, 0, 0, 0, 0, 0, 0, 0, 0]],
 'output': [[0, 0, 0, 0, 0, 0, 0, 0, 0, 0, 0, 0, 0, 0, 0, 2, 2, 2, 2, 0, 0, 0, 0, 0, 0, 0, 0]]}

### Task 8: 1d_move_2p
Rule: This task involves independently shifting two separate contiguous color blocks to new positions within the same array. Each block retains its original color and length, and the overall array length must remain unchanged.
The following constraints must be respected:
- There are exactly two non-zero color blocks in the input.
- Both blocks are relocated independently to different positions.
- The blocks must not overlap in the output.
- The structure, color, and size of each block must remain unchanged.
- All other positions not occupied by the two blocks are filled with zeros.
This task effectively repositions the blocks as if sliding them to new, predefined locations.
Example:
{'input':  [[0, 0, 0, 0, 0, 1, 1, 1, 1, 1, 1, 1, 0, 0, 0, 0]],
 'output': [[0, 0, 0, 0, 0, 0, 0, 1, 1, 1, 1, 1, 1, 1, 0, 0]]}
{'input':  [[0, 0, 6, 6, 6, 6, 6, 6, 0, 0, 0, 0, 0, 0, 0, 0]],
 'output': [[0, 0, 0, 0, 6, 6, 6, 6, 6, 6, 0, 0, 0, 0, 0, 0]]}

### Task 9: 1d_move_2p_dp
Rule: This task involves moving the **first contiguous color block (segment)** exactly **two positions toward the second block**, while the **second block remains stationary**. The following constraints must be respected:
The following constraints must be respected:
- Both blocks must retain their original **color** and **length**.
- The **overall length of the array must remain unchanged**.

- The direction of movement for the first block depends on its position relative to the second block (left or right).

Example valid transformations:
{'input': [[0, 0, 2, 2, 2, 2, 2, 2, 2, 2, 2, 2, 2, 2, 2, 2, 2, 2, 0, 0, 1, 0]],
'output': [[0, 0, 0, 0, 2, 2, 2, 2, 2, 2, 2, 2, 2, 2, 2, 2, 2, 2, 2, 2, 1, 0]]}
{'input': [[0, 0, 0, 0, 0, 0, 0, 0, 0, 4, 4, 4, 0, 0, 1, 0, 0, 0, 0, 0, 0]],
'output': [[0, 0, 0, 0, 0, 0, 0, 0, 0, 0, 0, 4, 4, 4, 1, 0, 0, 0, 0, 0, 0]]}

### Task 10: 1d_move_3p
Rule: This task involves moving a single contiguous color block exactly three positions to the right within a fixed-length 1D array. The color block retains its original color and length, and all other positions in the array are filled with zeros.
The following constraints must be respected:
- Only one non-zero color block is present in the input.
- The block must be shifted rightward by three cells.
- The overall length of the array remains unchanged.
- The block must not wrap around or exceed the array bounds.

Example:
{'input':  [[0, 9, 9, 9, 9, 9, 0, 0, 0]],
'output': [[0, 0, 0, 0, 9, 9, 9, 9, 9]]}
{'input':  [[8, 8, 8, 8, 0, 0, 0, 0, 0]],
'output': [[0, 0, 0, 8, 8, 8, 8, 0, 0]]}

### Task 11: 1d_move_dp
Rule: This task involves two distinct contiguous color blocks (segments). The **first block must move toward the second block**, shifting until it is directly adjacent to it. The **second block remains fixed in place**. The following constraints must be satisfied:
The following constraints must be respected:
- Both blocks must retain their original **color** and **length**.
- The **overall length of the array must remain unchanged**.
- The first block may move any number of positions right, but must stop once it becomes **immediately adjacent** to the second block.

Example valid transformations:
{'input': [[2, 2, 2, 2, 0, 0, 0, 0, 0, 0, 9]],
'output': [[0, 0, 0, 0, 0, 0, 2, 2, 2, 2, 9]]}
{'input': [[6, 6, 6, 6, 6, 0, 0, 0, 9, 0, 0, 0]],
'output': [[0, 0, 0, 6, 6, 6, 6, 6, 9, 0, 0, 0]]}

### Task 12: 1d_padded_fill
Rule: This task involves **filling background padding** around color blocks to create **uniform, continuous segments**. Whenever a color block is surrounded by padding (typically zeros), the padding within that segment should be replaced with the block's color, resulting in a solid, unbroken segment. Padding outside of these segments remains unchanged.
The following constraints must be respected:

- Only the padding **within** a segment (between repeated color values) is filled.
- The **color** and **relative location** of each segment must be preserved.
- The **overall length of the array remains unchanged**.

Example valid transformations:
{'input': [[0, 0, 4, 0, 0, 0, 0, 0, 0, 0, 4, 0, 0, 0, 0, 0, 0, 0, 4, 0, 0, 0, 0, 0, 0, 0, 4, 0, 0, 0, 0, 0, 0, 0, 4, 0, 0, 0, 0, 0, 0, 0, 0, 4, 0, 0, 0, 0, 0]],
'output': [[0, 0, 4, 4, 4, 4, 4, 4, 4, 4, 4, 4, 0, 0, 0, 0, 0, 0, 0, 4, 4, 4, 4, 4, 4, 4, 4, 4, 0, 0, 0, 0, 0, 0, 0, 0, 4, 4, 4, 4, 4, 4, 4, 4, 4, 4, 0, 0, 0, 0, 0]]}
{'input': [[0, 0, 0, 0, 0, 0, 0, 0, 0, 0, 0, 0, 0, 3, 0, 3, 0, 0, 0, 0, 0, 0, 0, 0, 0, 0, 0, 0, 0, 3, 0, 3, 0, 0, 0, 0, 0, 0, 0, 0, 0, 0, 0, 0, 0, 0, 0, 3, 0, 3, 0, 0]],
'output': [[0, 0, 0, 0, 0, 0, 0, 0, 0, 0, 0, 0, 0, 3, 3, 3, 0, 0, 0, 0, 0, 0, 0, 0, 0, 0, 0, 0, 0, 3, 3, 3, 0, 0, 0, 0, 0, 0, 0, 0, 0, 0, 0, 0, 0, 0, 0, 3, 3, 3, 0, 0]]}

### Task 13: 1d_pcopy_1c
Rule: This task involves duplicating the first contiguous color block and using its color to fill certain zero-surrounded positions later in the array.
- The following constraints must be respected:
- The first color block remains unchanged in its original position.
- Any subsequent color block or isolated color that is surrounded by zeros on both sides is replaced entirely with the color of the first block.
- Only segments that are fully bordered by zeros (i.e. 0, x, 0) are eligible for replacement.
- The overall length of the array remains unchanged.
This effectively propagates the first segment's color into other isolated regions.
Example:
{'input': [[0, 0, 1, 1, 1, 0, 0, 0, 0, 1, 0, 0, 0, 1, 0, 0, 0, 0, 0, 0, 0, 0, 0, 0, 0, 0, 0, 0, 0, 0, 0, 0]],
'output': [[0, 0, 1, 1, 1, 0, 0, 0, 1, 1, 1, 0, 1, 1, 1, 0, 0, 0, 0, 0, 0, 0, 0, 0, 0, 0, 0, 0, 0, 0, 0, 0]]}
{'input': [[0, 3, 3, 3, 0, 0, 0, 0, 3, 0, 0, 0, 0, 0, 0, 0, 0, 0, 0, 0, 0, 0, 0, 0, 0, 0, 0, 0, 0, 0, 0, 0]],
'output': [[0, 3, 3, 3, 0, 0, 0, 3, 3, 3, 0, 0, 0, 0, 0, 0, 0, 0, 0, 0, 0, 0, 0, 0, 0, 0, 0, 0, 0, 0, 0, 0]]}

### Task 14: 1d_pcopy_mc
Rule: This task involves identifying a single reference color block (always of length 3) at the start of the array, and then extending all subsequent isolated color cells (length 1) into full 3-cell blocks, using the same color.
The following constraints must be respected:
- The first color block (length 3) remains unchanged.
- Any subsequent single-color cells (i.e. 0, x, 0) are expanded into 3-cell blocks: [x, x, x].
- These expansions may overwrite adjacent zeros, but must not exceed array boundaries.
- The total length of the array remains the same.
This task effectively duplicates the structural pattern of the first block and applies it to isolated color values surrounded by zeros.
Example:
{'input': [[0, 0, 5, 5, 5, 0, 0, 0, 6, 0, 0, 0, 0, 0, 5, 0, 0, 0, 0, 0, 0, 0, 0, 0, 0, 0, 0, 0, 0, 0, 0, 0, 0]],
'output': [[0, 0, 5, 5, 5, 0, 0, 6, 6, 6, 0, 0, 0, 5, 5, 5, 0, 0, 0, 0, 0, 0, 0, 0, 0, 0, 0, 0, 0, 0, 0, 0, 0]]}
{'input': [[0, 7, 7, 7, 0, 0, 0, 9, 0, 0, 0, 8, 0, 0, 0, 0, 1, 0, 0, 0, 0, 0, 0, 0, 0, 0, 0, 0, 0, 0, 0, 0]],
'output': [[0, 7, 7, 7, 0, 0, 9, 9, 9, 0, 8, 8, 8, 0, 0, 1, 1, 1, 0, 0, 0, 0, 0, 0, 0, 0, 0, 0, 0, 0, 0, 0]]

}

### Task 15: 1d_recolor_cmp

Rule: This task involves recoloring color blocks based on their relative lengths. Specifically, blocks are recolored depending on whether they are the longest among all blocks, or not the longest. The recoloring rule may target only the longest blocks or only the non-longest blocks, depending on the specific task instance.
The following constraints must be respected:
- All color blocks retain their original positions and lengths.
- Zeros (0) are background and remain unchanged.
- Recoloring is applied in one of two possible modes:
- Mode A: Only the longest block(s) are recolored.
- Mode B: All non-longest blocks are recolored.
- The recolored blocks are assigned a new color (e.g., 3), while the rest remain unchanged.

This task highlights blocks based on comparative length criteria within the array.
Example:
{'input': [[0, 0, 9, 9, 9, 9, 0, 0, 9, 9, 0, 9, 9, 9, 0, 0, 0, 9, 9, 9, 9, 0]],
'output': [[0, 0, 3, 3, 3, 3, 0, 0, 9, 9, 0, 9, 9, 9, 0, 0, 0, 3, 3, 3, 3, 0]]}

{'input': [[0, 0, 9, 9, 9, 0, 9, 9, 0, 0, 0, 9, 9, 9, 9, 0, 0, 0, 9, 9, 9, 9, 9]],
'output': [[0, 0, 3, 3, 3, 0, 3, 3, 0, 0, 0, 9, 9, 9, 9, 0, 0, 0, 9, 9, 9, 9, 9]]}

### Task 16: 1d_recolor_cnt

Rule: This task involves recoloring contiguous color blocks based on their occurrence frequency, using a mapping learned from the training data. Specifically, the frequency (i.e., how many times a color block appears) determines its new color, according to the patterns observed in train.
The following constraints must be respected:
- All non-zero blocks retain their positions and lengths; only their colors change.
- The recoloring rule is not hardcoded, but learned from train:
  - Identify how often each color block appears in the input.
  - Find out what color those blocks became in the output.
- Build a mapping:
  → (original color, frequency) → new color
- In test data， apply the same mapping to perform recoloring.
- Background (0) values remain unchanged.

Example:
{ 'input': [[0, 0, 8, 8, 8, 0, 8, 0, 8, 8, 0, 0, 0, 8, 8, 0, 0, 8, 8, 8, 0, 0, 0, 0, 0]],
'output': [[0, 0, 1, 1, 1, 0, 7, 0, 3, 3, 0, 0, 0, 3, 3, 0, 0, 1, 1, 1, 0, 0, 0, 0, 0]]}
{ 'input': [[0, 8, 8, 8, 0, 0, 8, 8, 0, 0, 8, 0, 8, 8, 0, 8, 8, 0, 0, 0, 0, 0, 0, 0]],
'output': [[0, 1, 1, 1, 0, 0, 3, 3, 0, 0, 7, 0, 3, 3, 0, 3, 3, 0, 0, 0, 0, 0, 0, 0]]}

### Task 17: 1d_recolor_oe

Rule: This task involves recoloring contiguous non-zero color blocks based on the parity (odd or even) of their length, following a rule inferred from training data. The recoloring is applied only to blocks whose length parity matches a condition (odd or even), as determined from the train examples.

The following constraints must be respected:
- A block is defined as a contiguous sequence of the same non-zero color.
- The parity (odd/even) of each block's length is evaluated.
- From the training data, the model must infer:
  - Which type of length parity (odd or even) triggers recoloring.
  - What target color should be used for recoloring.
- Blocks not matching the target parity retain their original color.
- The position and length of each block remain unchanged.
- Zeros (0) are background and must not be recolored.

Example:
{'input':  [[0, 9, 9, 9, 0, 0, 9, 9, 0, 9, 9, 9, 9, 0]],
 'output': [[0, 3, 3, 3, 0, 0, 9, 9, 0, 9, 9, 9, 9, 0]]}
{'input':  [[0, 5, 5, 5, 5, 0, 0, 7, 0, 3, 3, 0]],
 'output': [[0, 1, 1, 1, 1, 0, 0, 7, 0, 1, 1, 0]]}

### Task 18: 1d_scale_dp
Rule: This task involves **scaling the first contiguous color block** by expanding it into the empty space (zeros) that lies between it and the second block. The second block remains fixed in its original position. The array length must remain unchanged.

The following constraints must be respected:
- Only the **first color block** is scaled (lengthened), using its original **color**.
- The **second color block remains unchanged** in both position and shape.
- The **space (zeros)** between the two blocks is partially or fully replaced with the first block's color to allow expansion.
- The **overall length of the array remains the same**.

Example valid transformations:
{'input': [[0, 3, 3, 3, 3, 3, 0, 0, 0, 5, 0, 0]],
 'output': [[0, 3, 3, 3, 3, 3, 3, 3, 3, 5, 0, 0]]}
{'input': [[0, 4, 4, 4, 4, 4, 0, 0, 5, 0, 0, 0]],
 'output': [[0, 4, 4, 4, 4, 4, 4, 4, 5, 0, 0, 0]]}

If a transformation matches one of the rules above, return the result in the following format:
******
<matched 18 types>
******
for example:
******
18) 1d_scale_dp
******

"""